\documentclass[journal]{IEEEtran}
\ifCLASSINFOpdf
\else
\fi

\usepackage{lineno,hyperref}
\usepackage{graphicx}
\usepackage{multirow}
\usepackage{amsmath}
\usepackage{float}
\usepackage{epstopdf}
\usepackage{tabularx}
\usepackage{subfig}
\usepackage{wasysym}
\usepackage{picins}
\begin{document}


%
\title{Attribute-guided Feature Learning Network for Vehicle Re-identification}
%
%
%


\author{\IEEEauthorblockN{Huibing Wang \IEEEauthorrefmark{2},
Jinjia Peng \IEEEauthorrefmark{2}, Dongyan Chen \IEEEauthorrefmark{2}
Guangqi Jiang \IEEEauthorrefmark{2}, Tongtong Zhao \IEEEauthorrefmark{3}
 and Xianping Fu \IEEEauthorrefmark{2}}


%
\thanks{\IEEEauthorrefmark{2}College of Information and Science Technology, Dalian Maritime University, Dalian, Liaoning, 116021, China, e-mail: \{huibing.wang,jinjiapeng,chendongyan,guangqi-j,fxp\}@dlmu.edu.cn}. \thanks{\IEEEauthorrefmark{3}zhaotongtong94@163.com}}
\markboth{Journal of \LaTeX\ Class Files,~Vol.~14, No.~8, August~2015}%
{Shell \MakeLowercase{\textit{et al.}}: Bare Demo of IEEEtran.cls for IEEE Journals}
%



\maketitle

\begin{abstract}
Vehicle re-identification (reID) plays an important role in the automatic analysis of the increasing urban surveillance videos, which has become a hot topic in recent years. However, it poses the critical but challenging problem that is caused by various viewpoints of vehicles, diversified illuminations and complicated environments. Till now, most existing vehicle reID approaches focus on learning metrics or ensemble to derive better representation, which are only take identity labels of vehicle into consideration. However, the attributes of vehicle that contain detailed descriptions are beneficial for training reID model. Hence, this paper proposes a novel Attribute-Guided Network (AGNet), which could learn global representation with the abundant attribute features in an end-to-end manner. Specially, an attribute-guided module is proposed in AGNet to generate the attribute mask which could inversely guide to select discriminative features for category classification. Besides that, in our proposed AGNet, an attribute-based label smoothing (ALS) loss is presented to better train the reID model, which can strength the distinct ability of vehicle reID model to regularize AGNet model according to the attributes. Comprehensive experimental results clearly demonstrate that our method achieves excellent performance on both VehicleID dataset and VeRi-776 dataset.

\end{abstract}


%
\IEEEpeerreviewmaketitle

\section{Introduction}

%
%
%
%
\IEEEPARstart{V}{ehicle}-related researches are of vital significance for intelligent transport, which have attracted more and more attention widely. Some progresses have been made in computer vision community, such as vehicle detection  \cite{hu2018sinet,ref_article2}, tracking \cite{tang2019cityflow,fang2019road} and classification \cite{wang2018multiview,ma2019fine}. Specially, vehicle reID is different from those tasks above and aiming to search a certain vehicle across large images captured from multiple non-overlapping cameras. Meanwhile, it could automatically carry out with less time consuming and manual labor by vehicle reID, which plays an important role in modern smart surveillance systems.

\begin{figure}[ht]
\centering
\includegraphics[width=8.9cm]{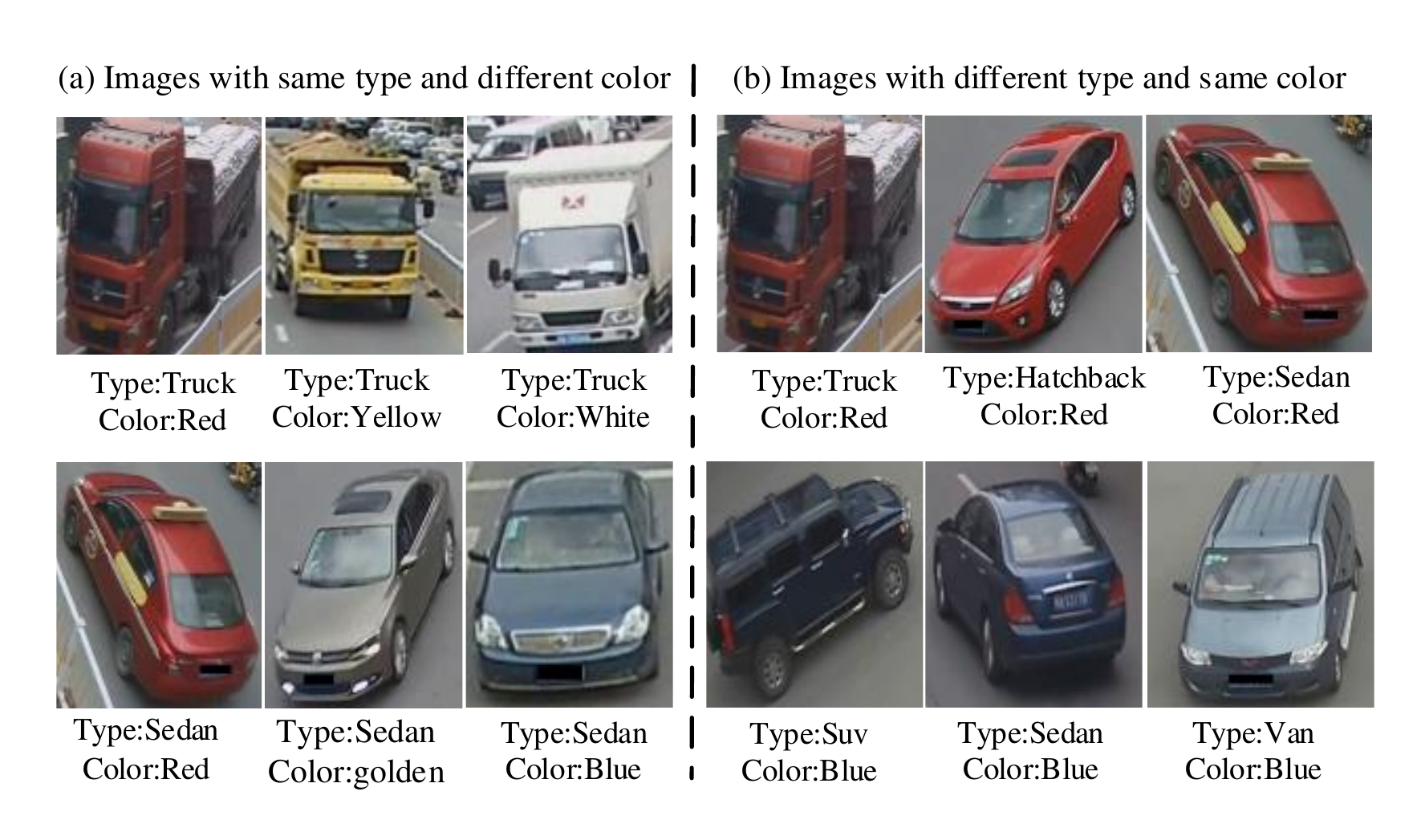}
\caption{Vehicle images with different attributes. (a) are the vehicle images with same type and different colors for each row. (b) are the vehicle images with same color and different types for each row.} \label{introduction}
\end{figure}

Despite recent progress in vehicle reID, in particular deep learning models have made some progress \cite{khorramshahi2019dual,he2019part,zhu2019vehicle,guo2018learning}, it still suffers from lots of difficulties caused by various viewpoints of vehicles, complicated environments and diversified illuminations, which makes a great difference in the visual appearance of vehicles. Different from other vision tasks \cite{wang2013clustering,wu2018cycle} , such as person ReID  \cite{wu2018and,wu20193,wu2019cross,li2017learning} and fine-grained \cite{yang2018learning,wu2018deep,ge2019weakly,wang2015unsupervised}, that can extract rich features from images with various poses and colors, vehicles usually have a few attributes that could be utilized to help extract distinctive features for similar vehicles. In particular, distinguishing vehicles that belong to the same or similar models can be more challenging.

To solve these problems, some existing methods focus on learning the spatio-temporal relationship between similar vehicles, such as \cite{wang2017orientation,shen2017learning}. However, the spatio-temporal information are not annotated in all existing datasets, which sets a limit to us for exploring it. Besides that, attributes that could provide rich information to learn the correlation among vehicles are important auxiliary signals in vehicle reID task. Hence, several approaches explore powerful features using the attribute of vehicles, such as \cite{liu2016deep,liu2018ram}. Attributes usually describe the high-level properties, which are discriminative for the vehicles. Taking images in Fig.\ref{introduction} as an example, which are selected from VeRi-776 dataset. Each row in the first part shows different vehicles, which have same model and different colors, while the images in the second column have the same color and different models. It is obvious that through the attribute of vehicles, it's easy to distinguish some vehicles. Hence, the attribute is one of the important cues for vehicle reID task, which could help make good use of some local details corresponding to the attribute.

\begin{figure*}[ht]
\centering
\includegraphics[width=14cm]{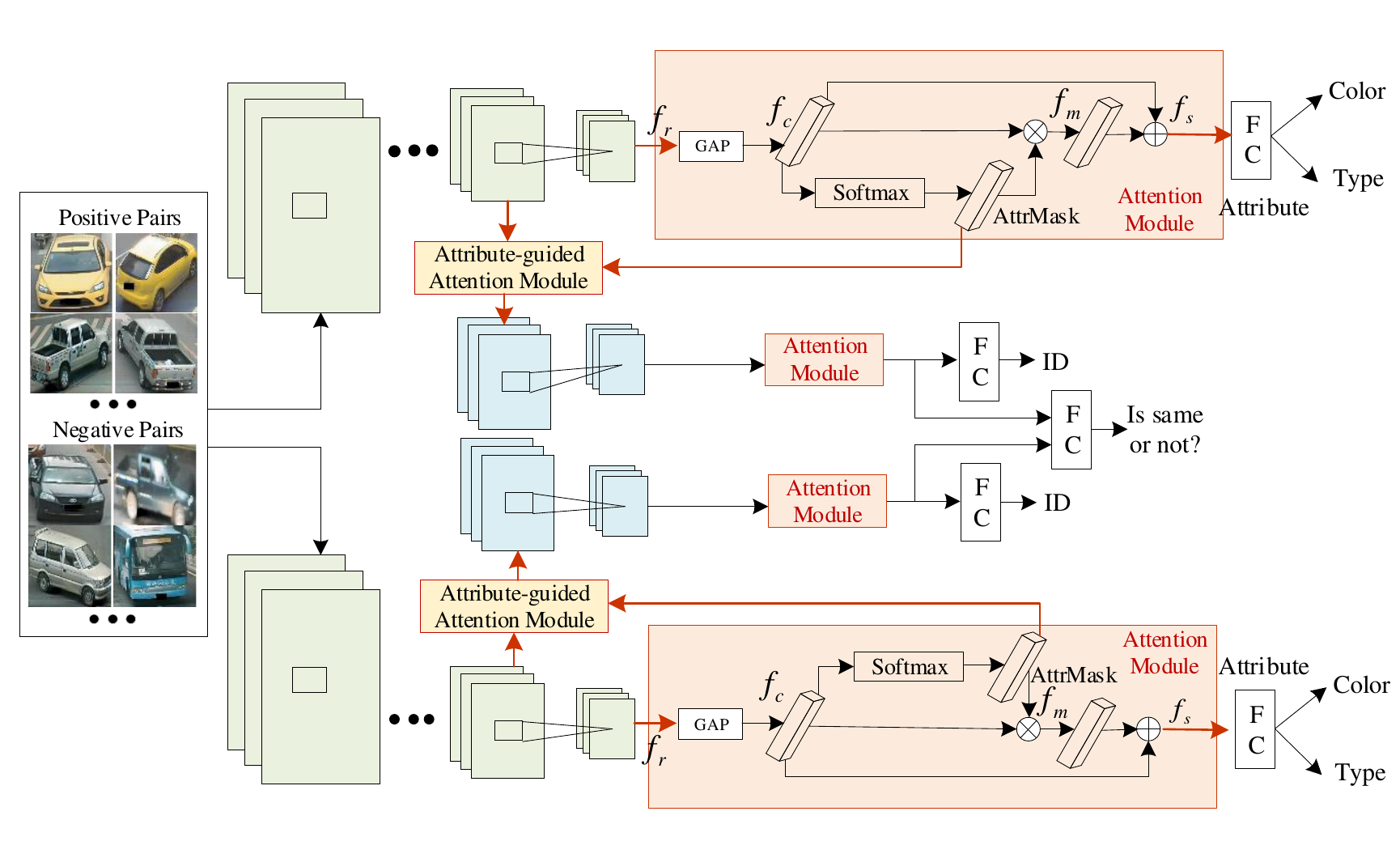}
\caption{Illustration of the AGNet network. AGNet is a dual-branch parallel structure. For one branch, after several resnet blocks, it is divided into two sub-branches that one is employed to predict attributes based on the image feature and another is for the category recognition. Specially, an attention module is proposed to generate the attribute mask, which helps the category branch to select better discriminative features for category recognition. Meanwhile, besides these recognition tasks, there is a verification task between two branches to improve the distinct ability of vehicle reID model.} \label{fig2}
\end{figure*}

Most existing methods train the attribute features by adding special single branch for different attributes, which neglects the relationship between the attributes and identity. Hence, in our paper, different from these methods, an Attribute-Guided Network (AGNet) is proposed to select details in category feature maps that are most relevant to intrinsic attributes, which improves the discriminative ability of vehicle reID model. Our intuition is that small attribute cues are usually crucial to distinguish different categories. Hence, besides training branches of attributes and category recognition respectively, an attention module is proposed in attribute branch to generate the mask that is inverse employed in the category branch, as shown in Fig.\ref{fig2}. Owning to the generated attribute mask, an attribute-guided attention module is designed in the category branch, which helps to refine category features simultaneously to choose the important attribute features for corresponding input vehicle images. Besides that, due to which the vehicles with same color or type are more similar than others. Hence, it is improper to treat all training sets equally. And the Attribute-based Label Smoothing (ALS) loss is proposed in this paper to better regular the AGNet for training more discriminative features. Our contributions are summarized as follows:

$\bullet$ An Attribute-Guided Network (AGNet) is proposed to simultaneously exploits complementarity between attributes and visual appearance. In AGNet, to better select discriminative features for vehicle reID, an attention module is proposed in attribute recognition branch to generate the mask, which guides the category classification branch to select local attribute features in category feature maps.

$\bullet$ The Attribute-based Label Smoothing (ALS) loss is proposed to assign different weights for the training sets, which takes the attributes of vehicles into consideration. The vehicles with same color or type mean the high possibility of the same vehicles and are assigned greater weights than others.

The rest of this paper is organized as follows. In section 2, we review and discuss related works. Section 3 illustrates the proposed method in detail. Experimental results and comparisons on two vehicle reID datasets are discussed in section 4, followed by conclusion in section 5.

\section{Related work}
In this section, existing vehicle reID works are reviewed. With the prosperity of deep learning, vehicle reID has achieved some progress in recent years. Broadly speaking, these approaches could be categorized into three classes, i.e., similarity learning, representation learning and spatio-temporal correlation learning.

A series of metric losses for deep feature embedding to achieve higher performance. In \cite{liu2016deep}, coupled cluster loss was proposed to pull the positive images closer and push those negative ones far away, which minimized intra-class distance and maximized inter-class distance to train the vehicle reID network. GST loss \cite{bai2018group} was introduced for CNNs to deal with intra-class variance in learning representation. Besides that, the mean-valued triplet loss was given to alleviate the negative impact of improper triplet sampling during training stage. MGR \cite{guo2019two} was presented to enhance the discrimination that not only between different vehicles but also different vehicle models, which further enhanced the discriminative ability of learned features.

Apart from designing losses, some methods attempt to identify vehicles based on the visual appearance. In \cite{zapletal2016vehicle}, full-fledged 3D bounding boxes vehicles were detected and then the color histograms and histograms of oriented gradients was used to extract features for vehicle reID. In \cite{zhao2019structural}, a ROIs-based vehicle reID method was proposed, which the ROIs' deep features were used as discriminative identifiers, encoding the structure information of a vehicle for reID task. VAMI \cite{zhou2018aware} transformed single-view feature into a global multi-view feature representation to better optimize the metric learning for training reID model. DHMVI \cite{zhou2018vehicle} utilized the spatially concatenated convnet and LSTM bi-directional loop to learn transformations across different viewpoints of vehicles, which could infer all viewpoints' information from the only one input view. Additionally, with the popular application of Generative Adversarial Networks (GAN) in person reID, some researchers adopt GAN in vehicle reID task. CV-GAN \cite{zhou2017cross} was first conducted to create the most likely image of other viewpoints to address the viewpoint variation problem. EALN \cite{lou2019embedding} was proposed to automatically generate hard negative samples in the specified embedding space, which improved the capability of the network for discriminating similar vehicles.

Besides, some approaches exploit spatial and temporal information for vehicle images to improve vehicle reID performance. PROVID \cite{liu2017provid} introduced the information of license plates, visual features and spatial-temporal relations with a progressive strategy to learn similarity scores between vehicle images. Siamese-Cnn+Path-LSTM \cite{shen2017learning} model was introduced to incorporate complex spatio-temoral information for regularizing the reID results. OIFE \cite{wang2017orientation} employed the log-normal distribution to model the spatio-temporal constrains in camera networks, which refined the retrieval results of vehicles.

\section{Attention-guided Network}
In this paper, we propose an end-to-end Attribute-guided feature learning network with ALS loss for vehicle reID. We firstly introduce the overview of the AGNet in section A. Then the detailed structure  of AGNet is illustrated in section B, C and D. Especially, the ALS loss is explained in the section D, which can help the AGNet to select better attribute feature.

\subsection{Architecture overview}
In this section, a novel Attribute-Guided Network (AGNet) is proposed, which simultaneously exploits complementarity between attributes and visual appearance. The pipeline of the proposed AGNet network is shown in Fig.\ref{fig2}. AGNet is a dual-branch parallel structure, which includes identification task and verification task. The images from the generation module are divided into positive and negative samples pairs as inputs for AGNet, which are captured by non-overlapping camera networks together with their corresponding vehicle ID. Images with the same vehicle IDs are positive sample pairs, otherwise, they are defined as negative sample pairs. The objective is to learn a discriminative representation for identifying the same vehicle and distinguishing different vehicles. Taking the upper branch as an example, given a pair of vehicle images, the AGNet extracts the feature representation $f_g$ by several resnet blocks \cite{He2015}. Subsequently, the branch is divided into two sub-branches, which one is employed to predict attributes based on the $f_g$ and another is for the category recognition. Specially, in the attribute sub-branch, an attention module is proposed to generate an attribute mask. The mask is then as additional cues to help the category branch select better discriminative features for category recognition. Meanwhile, in AGNet, besides the recognition task, it also contains the verification task between two parallel branches, which compares the two input features to judge that the vehicles are the same or not, and improve the distinct ability of vehicle reID model.

\subsection{Attribute recognition}
Attribute is a type of important auxiliary information for vehicle reID task. The same vehicles always have the same color and model. If vehicle images have different colors or models, they can't be the same vehicle. Hence, in our paper, an attribute recognition branch is designed in AGNet. As shown in Fig.\ref{fig2}, for one branch, the input image is fed into resnet blocks to output the feature map $f_{r}$ with the size of $2048\times 7\times 7$.  To focus on the meaningful parts of vehicle images and neglect the background when training the feature learning model, an attention module is proposed to generate distinct features. In the attention module, after a global average pooling layer, we employ the Softmax layer to re-weight the feature maps and generate the mask, which could be computed as:

\begin{equation}
M = Softmax(Conv(GAP(f_{r})))
\end{equation}
where the $Conv$ operator is $1\times 1$ convolution. The $M$ is the weight matrix, which contains cues of local attribute information. Hence, we call $M$ attribute mask (AttrMask). After obtaining the attention map $M$, the attended feature map could be calculated by $f_m = f_{a}\otimes M$. The operator $\otimes$ is performed in an element-wise product. Then the attended feature map $f_{m}$ is fed into the subsequent structure. At last, the prediction attribute classification is given by the fully connected layer with the cross-entropy loss that could be described as:

\begin{equation}
\ell_{attr}=\sum_{i=1}^k - p_i\log(q_i)
\end{equation}

Where $t$ is the target class and $q_i$ is the predicted probability. $k$ is the number of labels for attribute in the attribute recognition network. Specifically, $q_i$ could be calculated by Softmax. $p_i$ is the target probability. $p_i=0$ for all $i\ne y$ except $p_y=1$. There are vehicle color and vehicle type two attributes employed in this paper. Hence, there are two full connectional layers for different attributes.

\subsection{Category recognition}


\begin{figure}[ht]
\centering
\includegraphics[width=9cm]{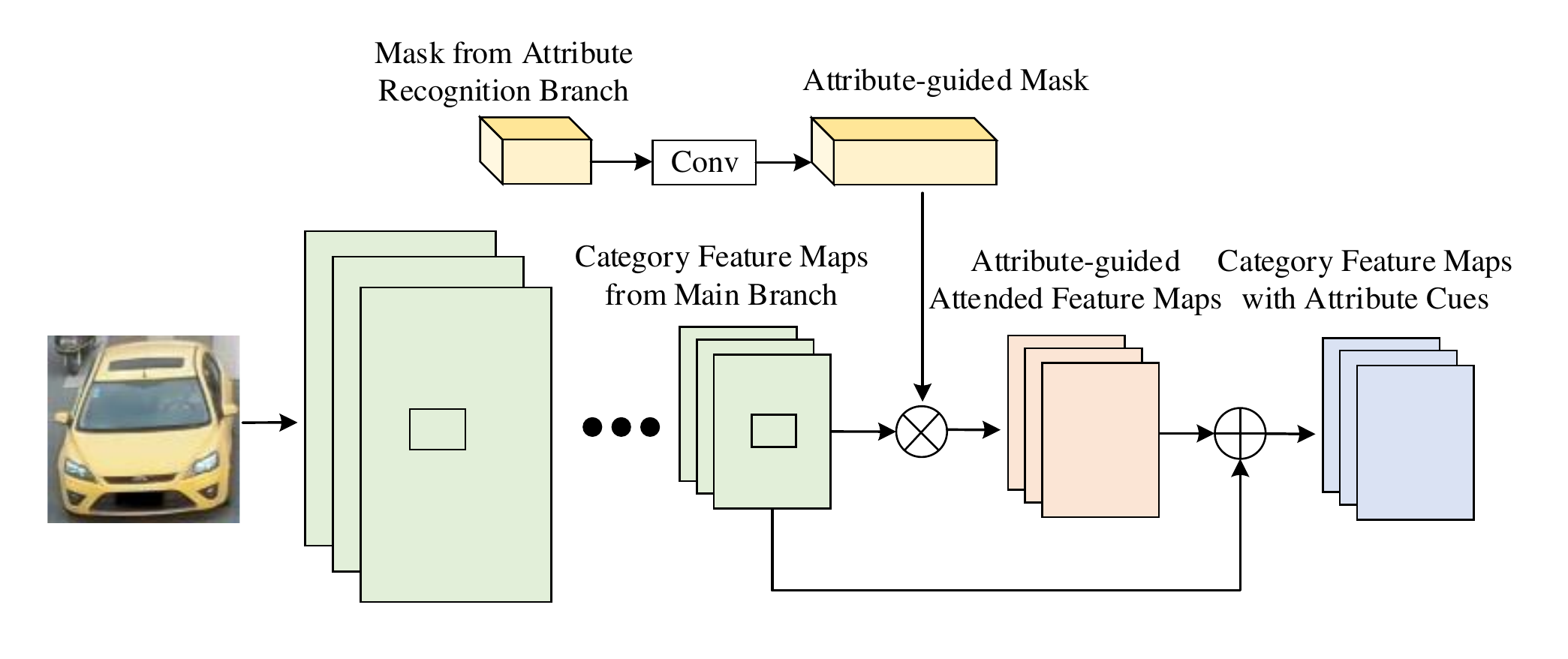}
\caption{Attribute-guided attention module. The attribute mask (AttrMask) is generated by the attribute recognition branch, which is then fed into the category recognition branch to produce an attention map for regional features with intrinsic attributes for the subsequent network} \label{fig4}
\end{figure}

To select regions in category feature maps which are most relevant to intrinsic attributes, we design an attribute-guided category recognition network to better generate discriminative features for category recognition task. Different from the attribute recognition, as shown in Fig. \ref{fig2}, the attribute mask (AttrMask) is generated by the attribute recognition branch, which is then fed into the category recognition branch to produce an attention map for details with intrinsic attributes for the subsequent network. As shown in Fig.\ref{fig4}, the attribute-guided attention weights are given in section 3.2 which could be described as $AttrMask$. The attribute features are multiplied by the attention weights and summed to produce the features $f_{attr}$

\begin{equation}
f_{attr} = f_c\otimes Conv(AttrMask)
\end{equation}
where $f_c$ is the category feature map. The operator $\otimes$ is performed in an element-wise product. Then the attended feature map $f_{attr}$ with attribute information is obtained. In order to fuse the attribute features and category features, a shortcut connection architecture is introduced to embed the input of the attention network directly to its output with an element-wise sum layer, which could be described as $f_{cs} = f_{c} + f_{attr}$. In this way, both the category-involved feature maps and the attribute-involved feature maps are combined to form features $f_{cs}$ and utilized as the input for the subsequent structure. Similar to the attribute branch, the cross-entropy loss is employed to train the category recognition, which could be described as:

\begin{equation}
\ell_{category}=\sum_{i=1}^k - p_i\log(q_i)
\end{equation}

Where $t$ is the target class and $q_i$ is the predicted probability. $k$ is the number of training identities int the dataset in the identification subnetwork.

During the test phrase, the final features are composed of features from category recognition branch and attribute classification branch, which could be described as follows:

\begin{equation}
f = [f_{attr},f_{category} \times (1-\alpha)]
\end{equation}
where $\alpha$ is the weight for features. $f_{category}$ is the feature from category recognition branch. $f_{attr}$ is from attribute recognition branch. The size of features from different branches are all $1\times 1\times 4096$.

\subsection{Verification network}

we add the Square Layer into the verification network. The Square Layer is denoted as $f_v=(f_1-f_2)^2$ ,where $f_1$ , $f_2$ are the 4096-dim embeddings and $f_v$ is the output tensor of the Square Layer. Then a convolutional layer and the ASL output function are added to embed the resulting tensor $f_v$ to a 2-dim vector. In the verification work, we treat it as a binary classification problem.

\begin{figure}[ht]
\centering
\includegraphics[width=8cm]{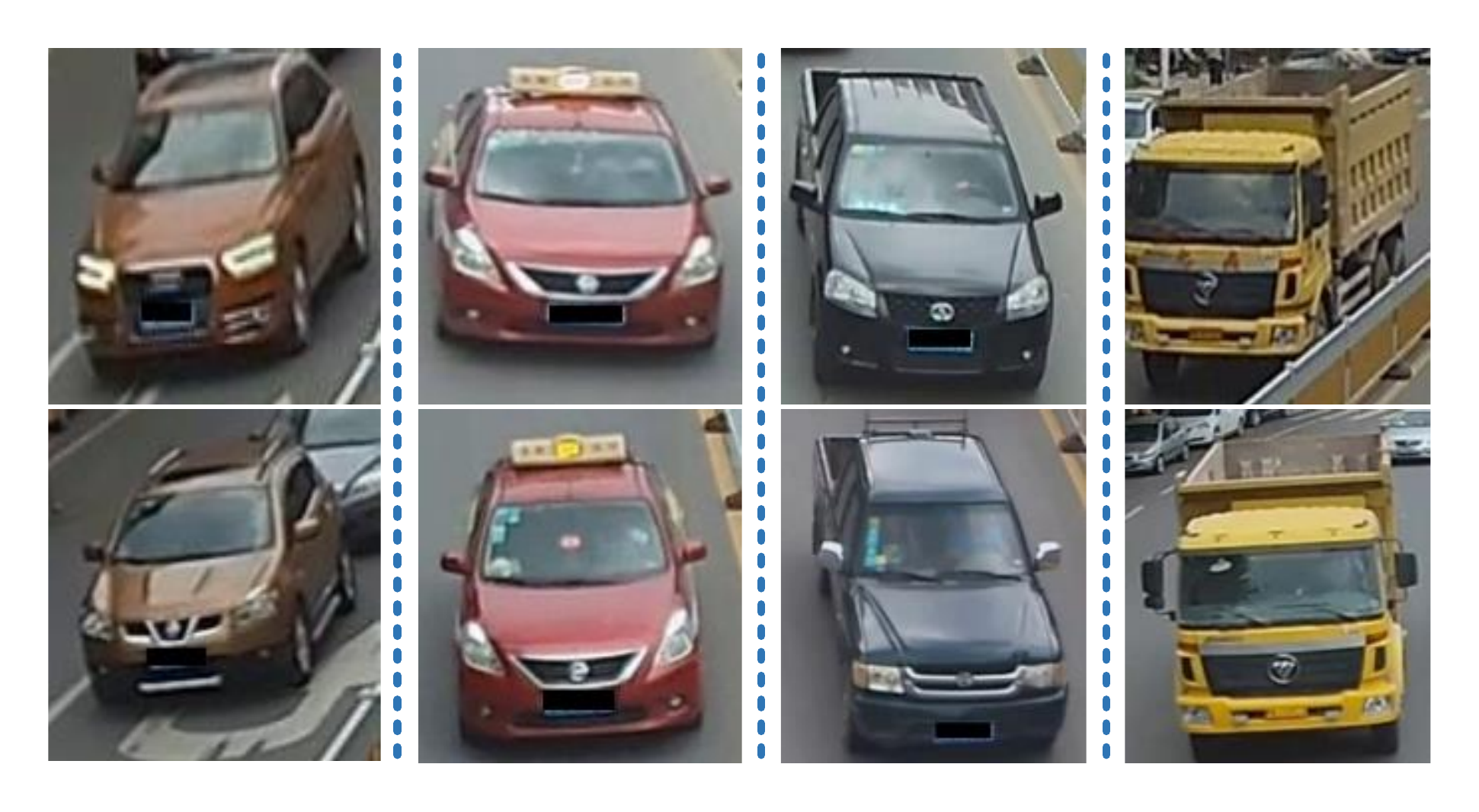}
\caption{The vehicle examples with different IDs but with the same vehicle color and type. It's difficult to discriminate these vehicles with similar appearance.} \label{figLoss}
\end{figure}

Traditionally, the cross-entropy loss is usually employed to train the reID model. However, The vehicles with the same type and color could have high degree of similarity, even though they have different IDs, as shown in Fig.\ref{figLoss}. Hence, these vehicles may be regard as hard training sets that help the network to effectively distinguish similarity vehicles. In the verification task, the ALS is proposed to regularize AGNet to learn better discriminative features, which could be described as follows:

\begin{equation}
\begin{split}
\ell_{verify} &= \ell_{CE} + \beta{\ell_{\gamma}} \\ & = \sum_{i=1}^k - p_i\log(q_i) + \sum_{i=1}^k - \varepsilon \times  p_i\log(\alpha+ q_i)
\end{split}
\end{equation}
where $\varepsilon$ is defined as:

\begin{equation}
  \varepsilon=
  \begin{cases}
   \theta, &\text{$attr_{1}=attr_{2},id_{1}\neq id_{2}$}\\
   1-\theta, &\text{$id_{1} = id_{2}$}\\
   0, &\text{if others,}\\
  \end{cases}
\end{equation}
where $id$ means the vehicle ID label and $attr$ represents the attribute of vehicles, such as vehicle type and vehicle color. $\theta$ is a hyper-parameter. If $\theta$ is set to 0, Eq.(6) is equivalent to Eq.(4). Besides, the $\alpha$ is a parameter for adjusting the possibility of the ground-truth. Compared with cross-entropy, the ALS is closer to reality by paying more additional attention to those vehicles with different vehicle IDs while are the same vehicle type and color.

\subsection{Training}
For the attribute-guided feature learning network, the overall objective function can be formulated as,

\begin{equation}
L_{\theta}=\lambda_1 \ell_{category}+\lambda_2 (\ell_{color} + \ell_{type}) + \lambda_3 \ell_{verify}
\end{equation}
where $\theta$ denotes the parameters in the deep model. $\ell_{category}$ is the category recognition loss. $\ell_{color}$ and $\ell_{type}$ are the attribute classification loss. $\ell_{verify}$ represents the loss of verification task. $\lambda_1$, $\lambda_2$ and $\lambda_3$ are the weights for corresponding loss. In our experiments, $\lambda_1$, $\lambda_2$ and $\lambda_3$ are set to 0.5, 0.5 and 1, respectively.

\section{Experiments}
\subsection{Datasets and evaluation metrics}
In this section, some detailed analyses are given to demonstrate the effectiveness of our method. And the proposed AGNet is evaluated utilizing the mean average precision (mAP)  \cite{lin2019improving} and the Cumulative Match Characteristic (CMC) curve, which are widely adopted in vehicle reID. First, we compare our AGNet with state-of-the-art methods. Then, the ablation studies are made to analyze each part in AGNet in detail. Various experiments are conducted on two popular vehicle reID datasets: VeRi-776 \cite{liu2017provid} and VehicleID \cite{liu2016deep}.

\subsubsection{Datasets.}

VeRi-776 \cite{liu2017provid} is a large urban surveillance vehicle dataset for reID, which contains more than 50,000 images of 776 vehicles with identity annotations, image timestamps, camera geo-locations, vehicle color and type information. In this paper, 37,781 images of 576 vehicles are utilized as the train set and 11,579 images of 200 vehicles are employed as the test set. A subset of 1,678 images in the test set generates the query set.

VehicleID \cite{liu2016deep} is a surveillance dataset from the real-world scenario, which contains 221,763 images corresponding to 26,267 vehicles in total. From the original testing data, four subsets are extracted, which contain 800, 1,600, 2,400 and 3,200 vehicles, and are searched in different scales. During the phrase of testing, an image is randomly selected from one identity to obtain a gallery set with 800 images, and then the remaining images are all employed as probe images. Three other test sets are processed in the same way.

\subsubsection{Evaluation metrics.}

In this paper, we use CMC curve and mAP to evaluate the overall performance for all test images. Each query image in a subset of test images is given for other test images, the average precision for each query $q$ is calculated by
\begin{equation}
AP(q)=\frac{\sum_{k=1}^nP(k)\times{rel(k)}}{N_{gt}}
\end{equation}

Where $P(k)$ denotes the precision at the $k_{th}$ position of the results. The $rel(k)$ is an indicator function equal to 1 if the $k_{th}$ result is correctly matched or zero otherwise. $n$ is the number of tests, and $N_{gt}$ is the ground truths. After experimenting for each query image, the $mAP$ will be calculated as follows:
\begin{equation}
mAP=\frac{\sum_{q=1}^QAP(q)}{Q}
\end{equation}

where $Q$  is the number of all queries. In this paper, the vehicle images with the same ID and camera number are considered to be junk images in our evaluation of results.

\subsection{Implementation details}

We implement the proposed vehicle reID model in the Matconvnet  \cite{vedaldi2015matconvnet} framework. The stochastic gradient descent is utilized with a momentum of $\mu=0.0005$ during the training procedure on both VeRi-776 and VehicleID. Due to the limit of the memory of GPU, the batch size is set to 32 on VeRi-776 and VehicleID. The learning rate of the first 50 epochs is set to 0.1, and the last 25 to 0.01. The mini-batch stochastic gradient descent (SGD) is adopted to update the parameters of the network. During the phrase of training, all images with full annotations are employed. However, there are only 10086 vehicles are annotated by attribute labels, which could be used in our proposed AGNet. Hence, there are 78956 images are employed in our experiments.

\begin{table}[htbp]
\renewcommand{\arraystretch}{1.3}
\centering
\caption{Experimental results on VeRi-776. The mAP (\%) and cumulative matching scores (\%) at rank 1, 5 are listed.}\label{tab1}
\begin{tabular}{p{3.7cm}|p{1.0cm}|p{1.0cm}|p{1.0cm}}
\hline
 \bfseries Method & \bfseries mAP & \bfseries Rank1 & \bfseries Rank5\\
\hline
\hline
LOMO \cite{liao2015person} &  9.64 & 25.33 & 46.48\\
DGD \cite{xiao2016learning} &  17.92 & 50.70 & 67.52\\
GoogLeNet \cite{yang2015large} &  17.81 & 52.12 & 66.79\\
FACT+Plate-SNN+STR \cite{liu2016deep2} &  27.77 & 61.64 & 78.78\\
NuFACT+Plate-REC \cite{liu2017provid} &  48.55 & 76.88 & 91.42\\
PROVID \cite{liu2017provid} &  53.42 & 81.56 & 95.11\\
Siamese-Visual \cite{shen2017learning} & 29.48 & 41.12 & 60.31\\
Siamese-Visual+STR \cite{shen2017learning} & 40.26 & 54.23 & 74.97\\
Siamese-CNN+Path-LSTM \cite{shen2017learning} & 58.27 & 83.49 & 90.04\\
OIFE+ST \cite{wang2017orientation} & 51.42 & 68.30 & 89.70\\
VAMI \cite{zhou2018aware} & 50.13 & 77.03 & 90.82\\
VAMI+ST \cite{zhou2018aware} & 61.32 & 85.92 & 91.84\\
\hline
\hline
AGNet-ASL &  66.32 & 90.90 & 96.20\\
AGNet-ASL+STR & 71.59 & 95.61 & 96.56\\
\hline
\end{tabular}
\end{table}

\begin{table*}[htbp]
\renewcommand{\arraystretch}{1.3}
\centering
\caption{Experimental results on VehicleID. The mAP (\%) and cumulative matching scores (\%) at Rank 1, 5 are listed.}\label{tab2}
\begin{tabular}{p{2cm}|p{1cm}|p{1cm}|p{1cm}|p{1cm}|p{1cm}|p{1cm}|p{1cm}|p{1cm}|p{1cm}}
\hline
 \multirow{2}*{ \bfseries  Method} & \multicolumn{3}{c|}{ \bfseries Test size = 800} & \multicolumn{3}{c|}{ \bfseries Test size = 1600}& \multicolumn{3}{c}{ \bfseries Test size = 2400} \\ \cline{2-10} &  mAP & Rank1 & Rank5 & mAP & Rank1	& Rank5 & mAP & Rank1 & Rank5 \\
\hline
\hline
BOW-SIFT \cite{liu2016deep2}   & -  & 2.81  & 4.23 &- & 3.11 & 5.22 & -	& 2.11	& 3.76\\
LOMO \cite{liao2015person}  & -  & 19.76  & 32.14 &- & 18.95 & 29.46 & -	& 15.26	& 25.63\\
DGD \cite{xiao2016learning}  & -  & 44.80  & 66.28 &- & 40.25 & 65.31 & -	& 37.33	& 57.82\\
VGG+T \cite{liu2016deep}    & -  & 40.4  & 61.7 &- & 35.4 & 54.6 & -	& 31.9	& 50.3\\
VGG+CCL \cite{liu2016deep}  & -	&43.6	&64.2	&-	&42.8	&66.8	&-	&32.9	&53.3\\
Mixed DC \cite{liu2016deep}   & -	&49.0	&73.5	&-	&42.8	&66.8	&-	&38.2&	61.6\\
FACT  \cite{liu2017provid}  & -	&49.53	&67.96	&-	&44.63	&64.19	&-	&39.91&	60.49\\
NuFACT \cite{liu2017provid}  & -	&48.90	&69.51	&-	&43.64	&65.34	&-	&38.63&	60.72\\
OIFE \cite{wang2017orientation}   &-	&-&	-	&-	&-	&-	&-	&67.0	&82.9\\
VAMI \cite{zhou2018aware} &-	&63.12&	83.25	&-	&52.87	&75.12	&-	&47.34	&70.29\\
TAMR \cite{guo2019two}    &67.64	&66.02 &79.71	&63.69	&62.90	&76.80	&60.97	&59.69	&73.87\\
\hline
\hline
AGNet-ASL    &74.05 &71.15	&83.78 &72.08 &69.23 &81.41	&69.66 &65.74 &78.28\\
\hline
\end{tabular}
\end{table*}

\subsection{Comparison with the state-of-the-art methods}

\subsubsection{Comparison on VeRi-776} The results of the proposed method is compared with state-of-the-art methods on VeRi-776 dataset in Tables \ref{tab1} \ref{tab2}, which includes: (1) LOMO \cite{liao2015person}; (2) DGD \cite{xiao2016learning}; (3) GoogLeNet \cite{yang2015large} (4) FACT+Plate-SNN+STR \cite{liu2016deep2}; (5) NuFACT+Plate-REC \cite{liu2017provid}; (6) PROVID \cite{liu2017provid}; (7) Siamese-Visual \cite{shen2017learning}; (8) Siamese-Visual+STR \cite{shen2017learning}; (9) Siamese-CNN+Path-LSTM \cite{shen2017learning}; (10) OIFE+ST \cite{wang2017orientation}; (11) VAMI \cite{zhou2018aware}; (12) VAMI+ST \cite{zhou2018aware}. From the Tables \ref{tab1} \ref{tab2}, it should be noted that the proposed method achieves the best performance among the compared with methods with rank-1 = 90.90\%, mAP = 66.32\% on VeRi-776, which acquires the highest mAP and rank-1 among all methods under comparisons. More details are analyzed as follows.

Firstly, the proposed AGNet obtains much better performance than those hand-crafted feature representation methods, such as LOMO \cite{liao2015person} and DGD \cite{xiao2016learning}, which achieves 56.68 and 48.40 points in mAP improvements, respectively. This verifies that the features obtained from deep model are more robust than the hand-crafted feature that are severely affected by the complicated environment.

Secondly, spatio-temporal information is one of the most important cues for vehicle reID. Compared with deep learning methods that exploit vehicle reID task with spatio-temporal information, such as FACT+Plate-SNN+STR \cite{liu2016deep2}, PROVID \cite{liu2017provid}, Siamese-Visual+STR \cite{shen2017learning}, Siamese-CNN+Path-LSTM \cite{shen2017learning}, OIFE+ST \cite{wang2017orientation} and VAMI+ST \cite{zhou2018aware}, the proposed AGNet has higher mAP, rank-1 and rank-5 than them, which demonstrates that our AGNet could extract more discriminative features without other information besides the vehicle images.

Thirdly, compared with those single modal deep learning based methods, which trains the reID model without the spatio-temporal information, the proposed AGNet shows a larger accuracy improvement. Specifically, the best single modal deep learning method VAMI \cite{zhou2018aware} is also lower than the proposed AGNet in mAP, rank-1 and rank-5, which significantly shows the effectiveness of the proposed AGNet.

\subsubsection{Comparison on VehicleID} There are 9 methods are compared with our proposed method, which are (1) LOMO \cite{liao2015person}; (2) DGD \cite{xiao2016learning}; (3) VGG+T \cite{liu2016deep}; (4) VGG+CCL \cite{liu2016deep}; (5) Mixed DC \cite{liu2016deep}; (6) FACT \cite{liu2017provid}; (6) NuFACT \cite{liu2017provid}; (7) OIFE \cite{wang2017orientation}; (8) VAMI \cite{zhou2018aware}; (9) TAMR \cite{guo2019two}. Firstly, it can be observed that deep learning based methods obviously outperform traditional methods. And compared with traditional methods LOMO \cite{liao2015person} and DGD \cite{xiao2016learning}, the proposed method AGNet has 68.34\% and 51.39\% gains on the test size with 800 vehicles, respectively. Secondly, Different VeRi-776, there is no spatio-temporal labels in VehicleID. Hence, there are no methods that consider the spatio-temporal information. All compared methods utilize the information only from vehicle images. The proposed AGNet outperforms all deep learning based methods under comparison on the test sets with different sizes on VehicleID, which obtains 71.15\%, 69.23\%, 65.74\% in rank-1, respectively.  And this also shows that our proposed AGNet could generate more distinct features for different vehicle reID datasets.

\begin{table}[htbp]
\renewcommand{\arraystretch}{1.3}
\centering
\caption{Performance of features fusion on VeRi-776. The mAP (\%) and cumulative matching scores (\%) at Rank 1, 5 are listed.}\label{tab3}
\begin{tabular}{p{2cm}|p{1.0cm}|p{1.0cm}|p{1.0cm}}
\hline
\bfseries Descriptor & \bfseries  mAP & \bfseries Rank1 & \bfseries Rank5\\
\hline
\hline
Only-ID &  57.82 & 87.24 & 93.32\\
\hline
AGNet-CE-Attr &  44.09 & 81.34 & 89.74\\
AGNet-CE-ID &  64.09 & 89.78 & 94.79\\
AGNet-CE-All &  64.78 & 89.86 & 95.17\\
\hline
AGNet-ASL-Attr & 43.68 & 80.15 & 90.22\\
AGNet-ASL-ID &  62.61 & 89.69 & 94.75\\
AGNet-ASL-All &  66.32 & 90.90 & 96.20\\
\hline
\end{tabular}
\end{table}

\begin{figure}[ht]
\centering
\includegraphics[width=7cm]{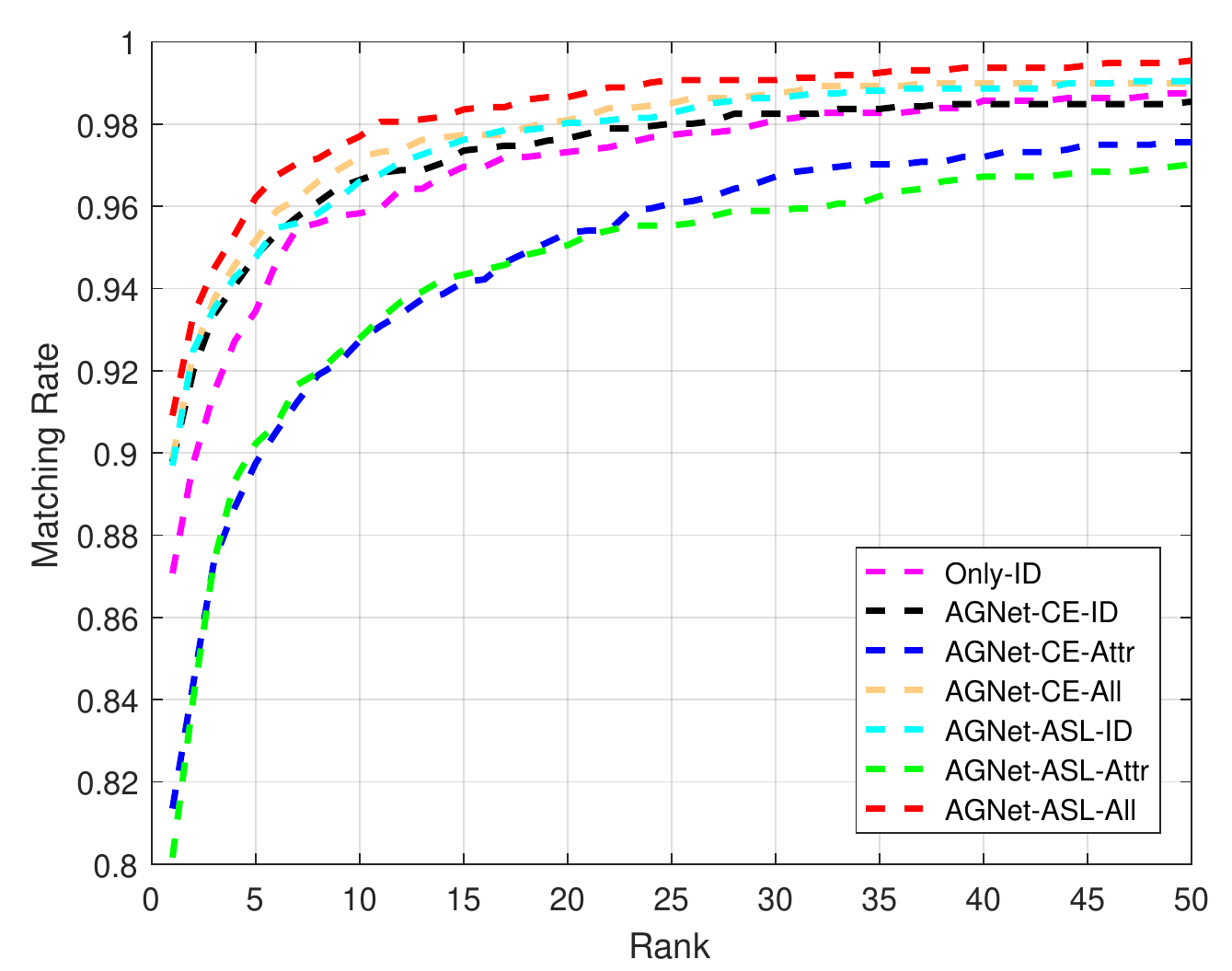}
\caption{ The CMC results of different methods on VeRi-776. } \label{fig7}
\end{figure}

\begin{table*}[htbp]
\renewcommand{\arraystretch}{1.3}
\centering
\caption{Performance of features fusion on VehicleID. The mAP (\%) and cumulative matching scores (\%) at Rank 1, 5 are listed.}\label{tab4}
\begin{tabular}{p{2.2cm}|p{0.8cm}|p{0.8cm}|p{0.8cm}|p{0.8cm}|p{0.8cm}|p{0.8cm}|p{0.8cm}|p{0.8cm}|p{0.8cm}|p{0.8cm}|p{0.8cm}|p{0.8cm}}
\hline
\multirow{2}*{ \bfseries Descriptor} & \multicolumn{3}{c|}{ \bfseries Test size = 800} & \multicolumn{3}{c|}{ \bfseries Test size = 1600}& \multicolumn{3}{c|}{ \bfseries Test size = 2400} & \multicolumn{3}{c}{ \bfseries Test size = 3200} \\
\cline{2-13} & mAP & Rank1 & Rank5 & mAP & Rank1 & Rank5 & mAP & Rank1 & Rank5 & mAP & Rank1 & Rank5 \\

\hline
 Only-ID          &70.59 &67.56 &80.61 &68.97 &65.87 &79.36 &65.06 &61.88 &75.26 &63.49 &60.62 &72.60\\
\hline
 AGNet-CE-Attr    &67.58 &63.60 &81.01 &62.47 &58.39 &76.30	&59.17 &54.95 &73.37 &56.60 &52.71 &69.23\\
 AGNet-CE-ID      &71.54 &68.73 &80.90 &69.22 &66.41 &78.20	&65.67 &62.74 &75.15 &64.34 &61.69 &72.71\\
 AGNet-CE-All     &72.20 &69.19 &82.48 &69.52 &66.52 &79.25	&67.50 &63.99 &76.86 &65.24 &62.51 &73.97\\
\hline
 AGNet-ASL-Attr   &66.77 &62.69	&81.29 &63.22 &59.20 &76.78	&59.59 &55.53 &73.06 &57.09 &53.34 &69.29\\
 AGNet-ASL-ID     &72.39 &69.66	&81.55 &70.68 &67.99 &79.40	&66.95 &64.09 &76.46 &65.63 &63.11 &73.42\\
 AGNet-ASL-All    &74.05 &71.15	&83.78 &72.08 &69.23 &81.41	&69.66 &65.74 &78.28 &67.02 &64.40 &75.21\\
\hline
\end{tabular}
\end{table*}

\begin{figure*}[ht]
\centerline{
\subfloat[Test size=800]{\includegraphics[width=1.8in,height=1.4in]{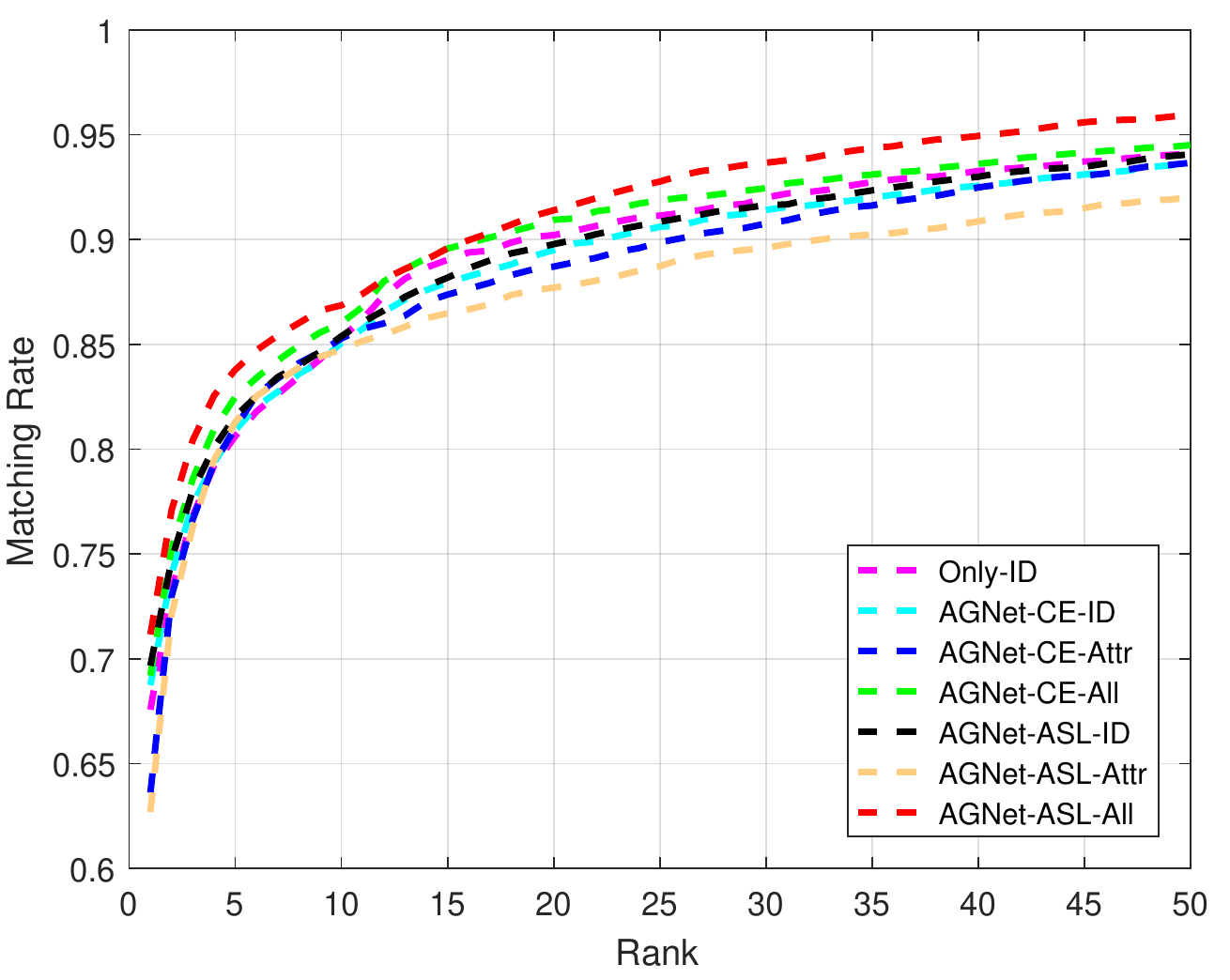}}
\subfloat[Test size=1600]{\includegraphics[width=1.8in,height=1.41in]{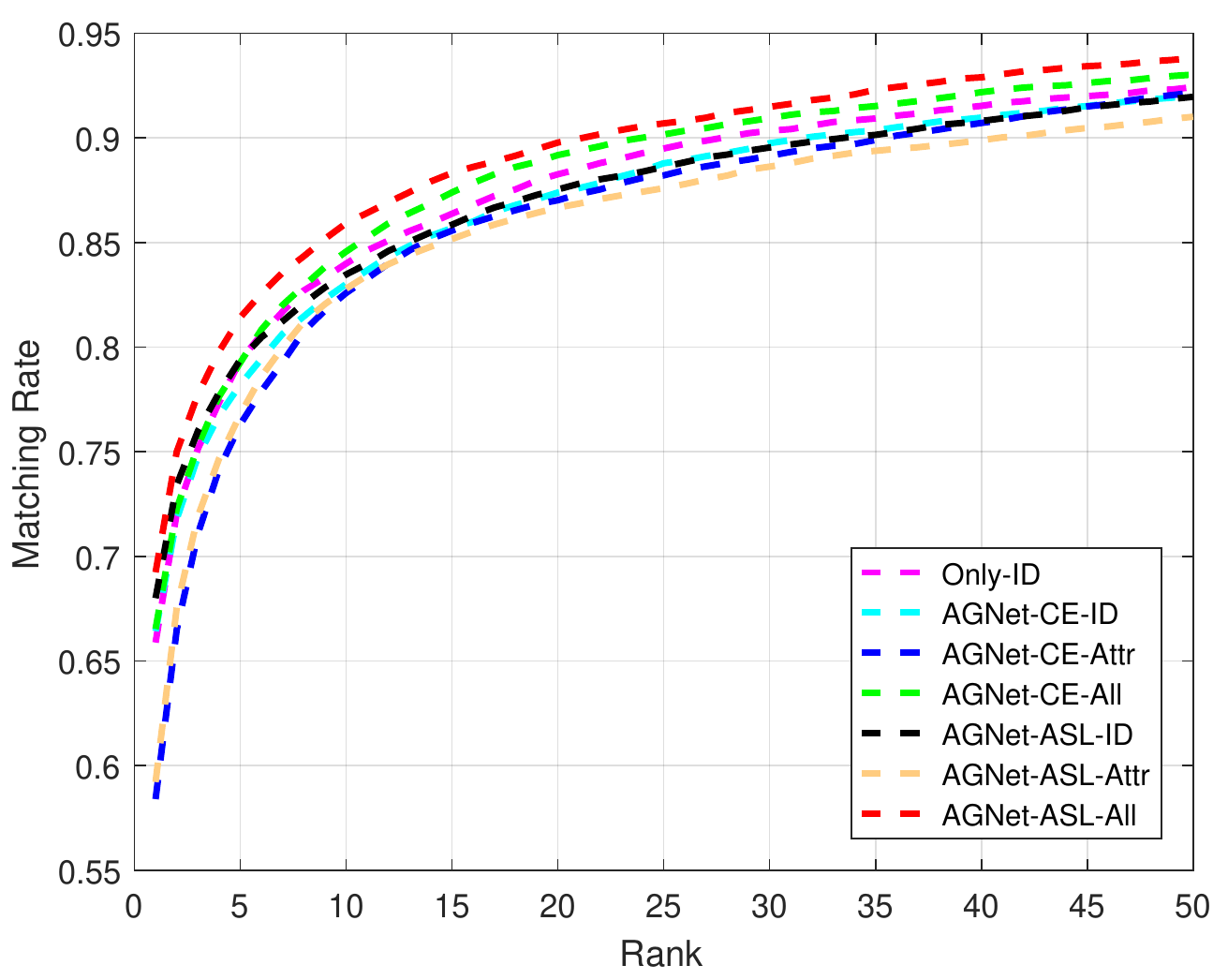}}
\subfloat[Test size=2400]{\includegraphics[width=1.8in,height=1.4in]{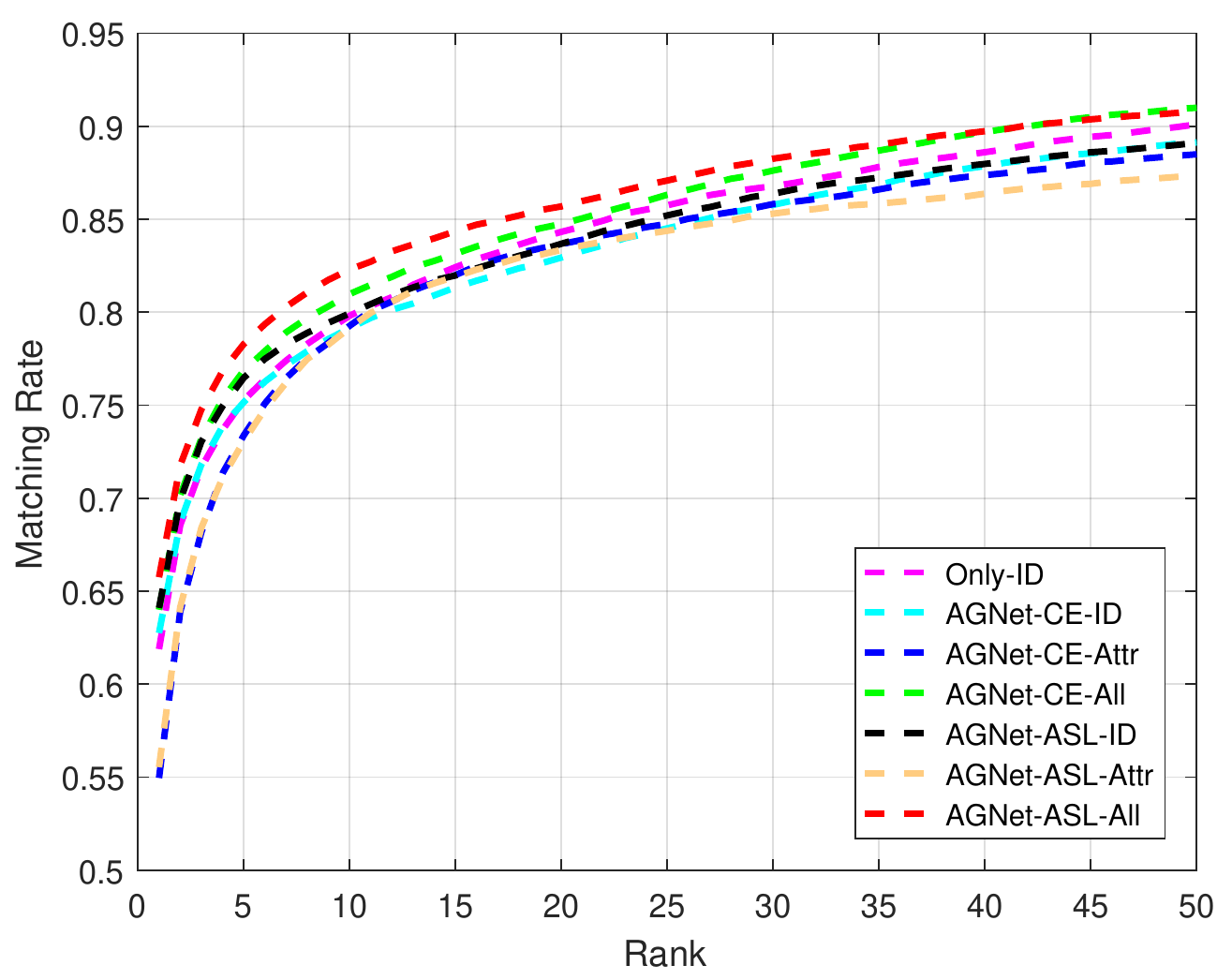}}
\subfloat[Test size=3200]{\includegraphics[width=1.8in,height=1.4in]{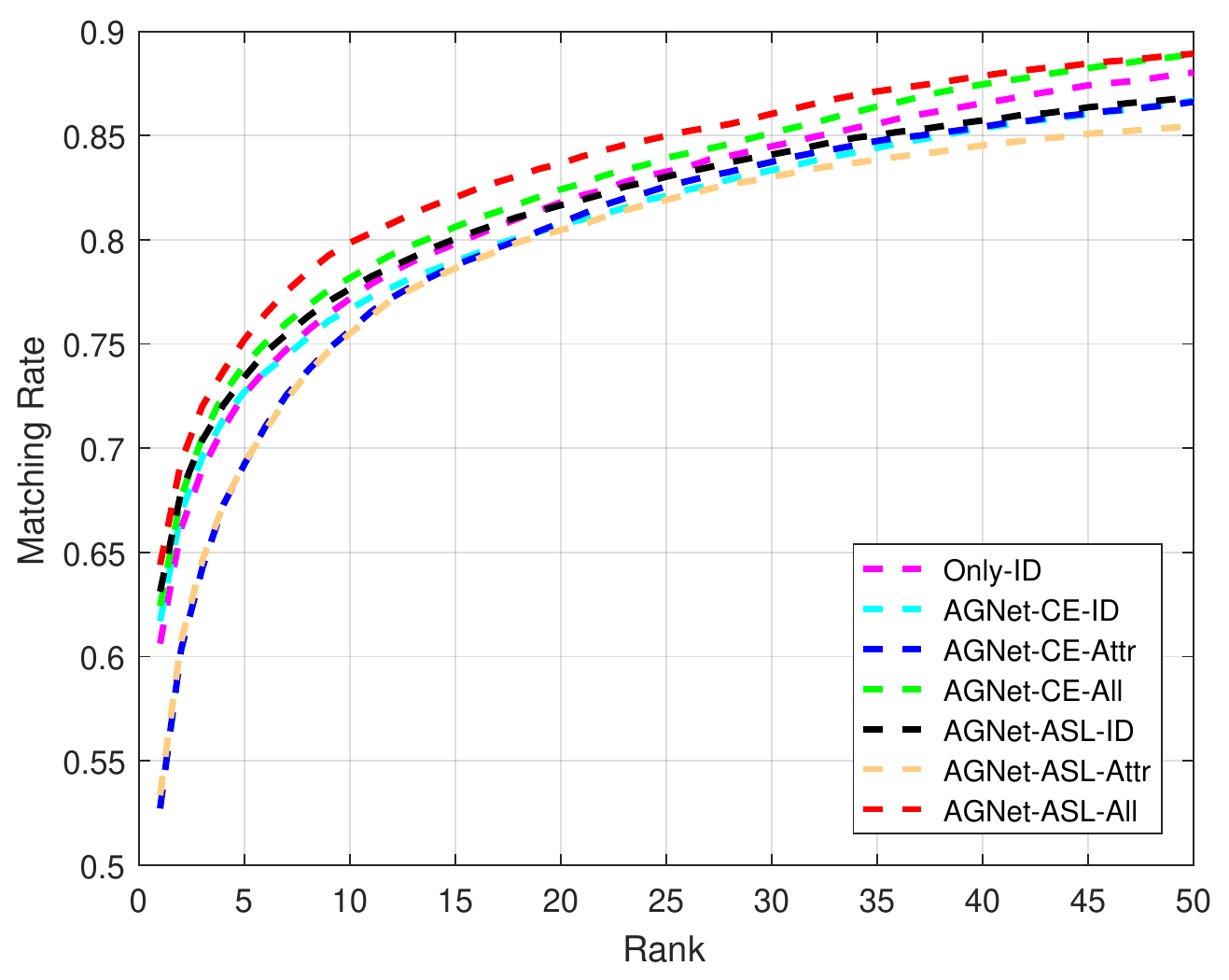}}
}
\caption{The CMC curves of different methods on VehicleID.} \label{fig8}
\end{figure*}

\begin{figure}[ht]
\centering
\includegraphics[width=8cm]{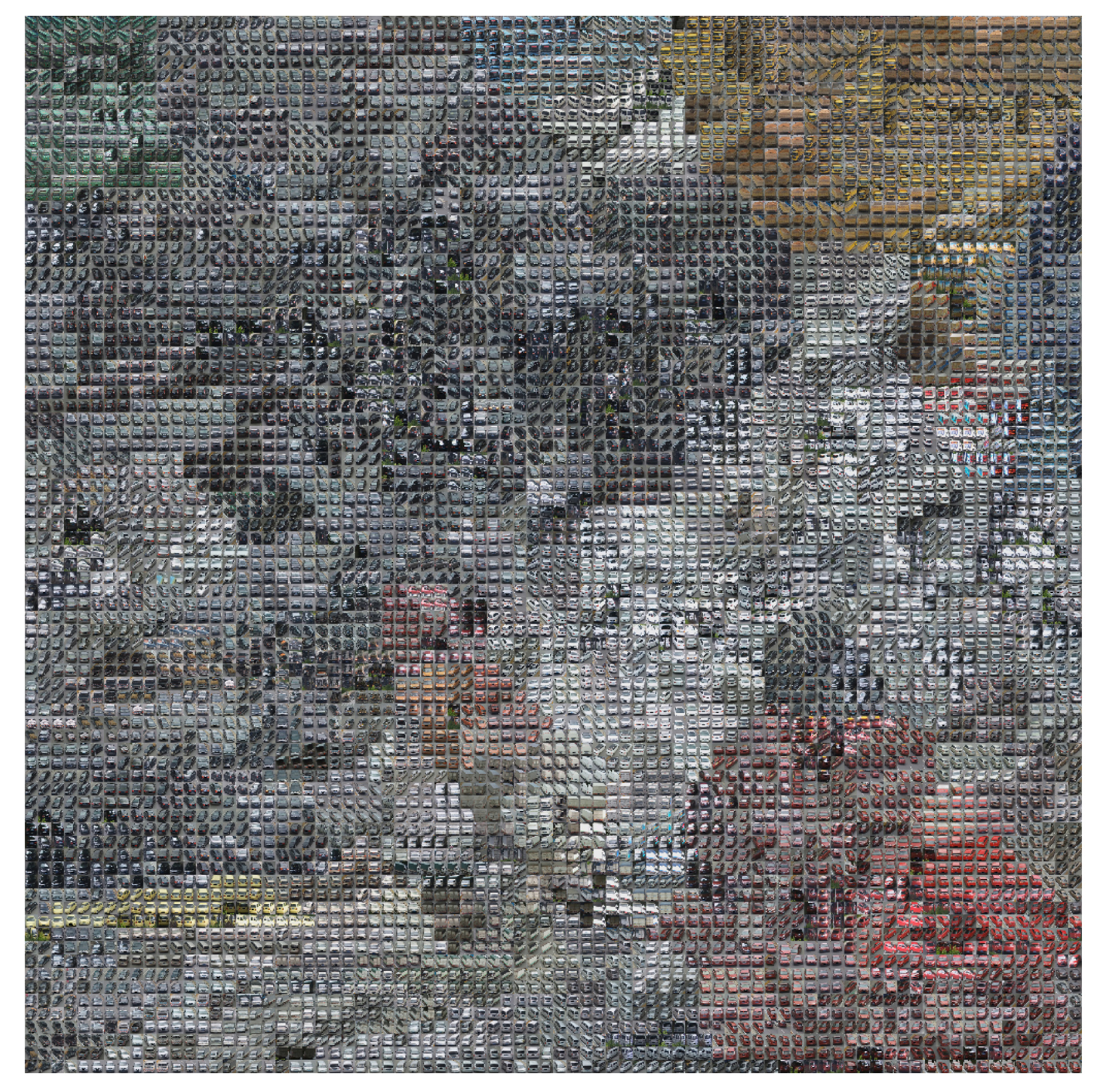}
\caption{Illustration of the features obtained from our proposed network on a test set of VeRi-776 with t-SNE. Best viewed in color. In addition to vehicle images with the same ID, the images with similar colors or types are more easily concentrated together.} \label{tsne}
\end{figure}

\subsection{Evaluation of proposed method}

To verify the effectiveness of the proposed method, some ablation experiments are conducted. The comparison results on VeRi-776 and VehicleID are presented in Table \ref{tab3} and Table \ref{tab4}. Fig. \ref{fig7} and Fig. \ref{fig8} show the CMC curves of the compared methods.

Firstly, our descriptor is learned by multiple branches in the proposed network. Every branch is trained with sharing parameters in part of convolutional layers. We thus compare different features to test the effectiveness of our descriptor. ``AGNet-ASL-All'' is our proposed method that combines all features for reID task. ``AGNet-ASL-ID'' and ``AGNet-ASL-Attr'' denote the features are extracted by category branch and attribute branch, respectively. Different from ``AGNet-CE-All'', ``AGNet-CE-All'' means that the training network has the same structure with ``AGNet-ASL-All'' except the training loss. ``AGNet-ASL-All'' is trained with the proposed ASL while ``AGNet-CE-All'' is CE loss. ``AGNet-CE-Attr'' and ``AGNet-CE-ID'' are similar with ``AGNet-ASL-ID'' and ``AGNet-ASL-Attr''. From Table \ref{tab3} and Table \ref{tab4}, it can be observed that ``AGNet-ASL-ID'' achieves higher mAP and rank-1 than ``AGNet-ASL-Attr''. This is because that ``AGNet-ASL-ID'' obtained from category branch which is trained with identity information has more details than the features from attribute branch. Besides that, it is worth noting that ``AGNet-ASL-All'' has 3.71\% improvements in mAP than the ``AGNet-ASL-ID'' on VeRi-776. And it also has 22.64\% increases than the `AGNet-ASL-Attr''. The similar improvements could be observed from the comparison of ``AGNet-CE-All'', ``AGNet-ASL-ID'' and ``AGNet-ASL-Attr''. These could demonstrate that the attribute branch learns some distinctive features which are helpful for vehicle reID task.

Secondly, to demonstrate the effectiveness of the proposed attribute-guided attention model, the ``Only-ID'' is compared, which trains the reID model only with the ID labels with the attention structure. It is observed that, compared with ``Only-ID'', our proposed ``AGNet-ASL-All'' achieves 8.5\% improvements in mAP on VeRi-776. For VehicleID, it has 3.94\%, 3.36\%, 3.86\%, 3.78\% gains in rank-1 on test sets with different size.

\begin{figure*}[htbp]
\centerline{
\subfloat[VeRi-776]{\includegraphics[width=15cm]{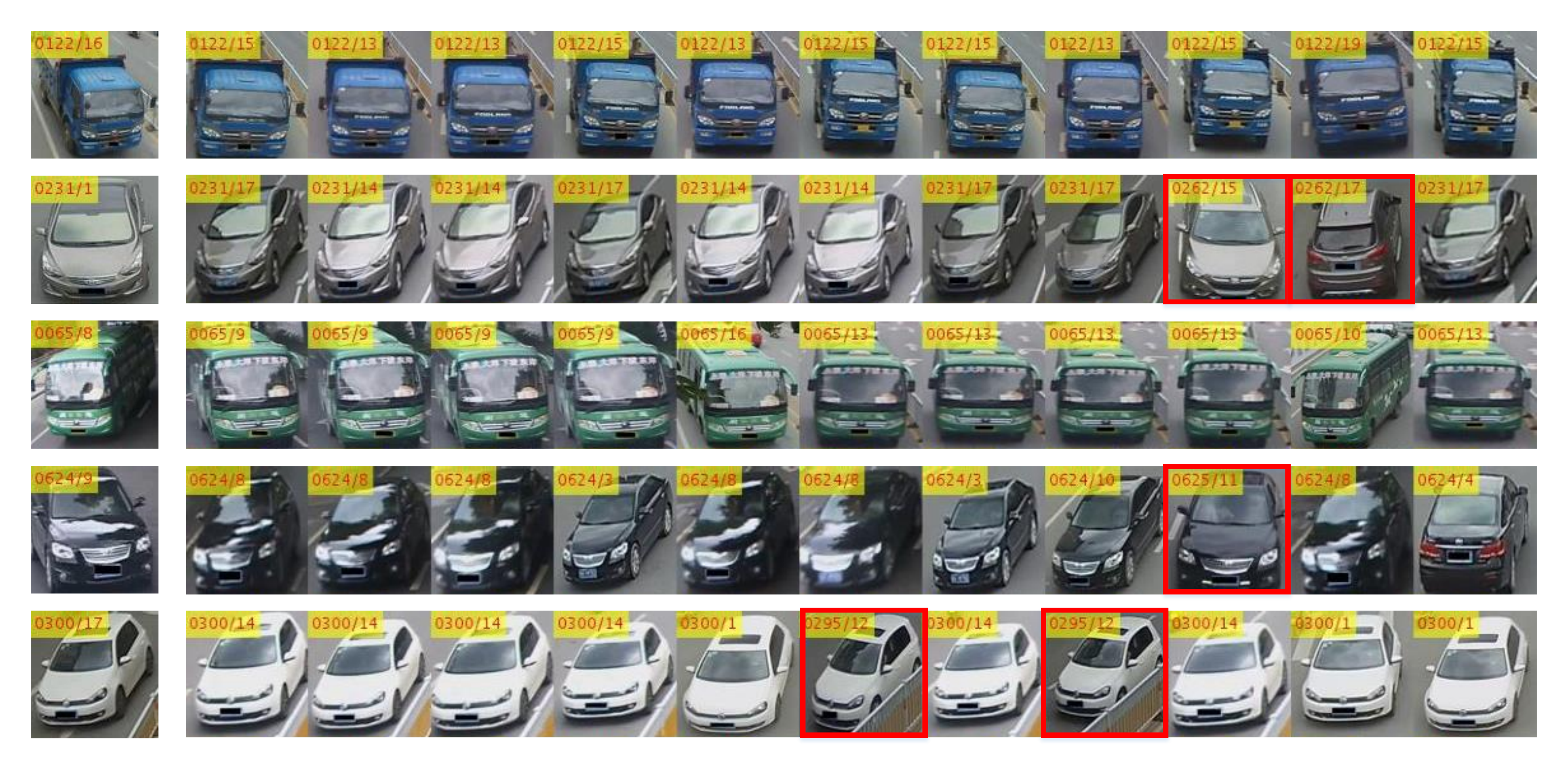}}
}
\centerline{
\subfloat[VehicleID]{\includegraphics[width=15cm]{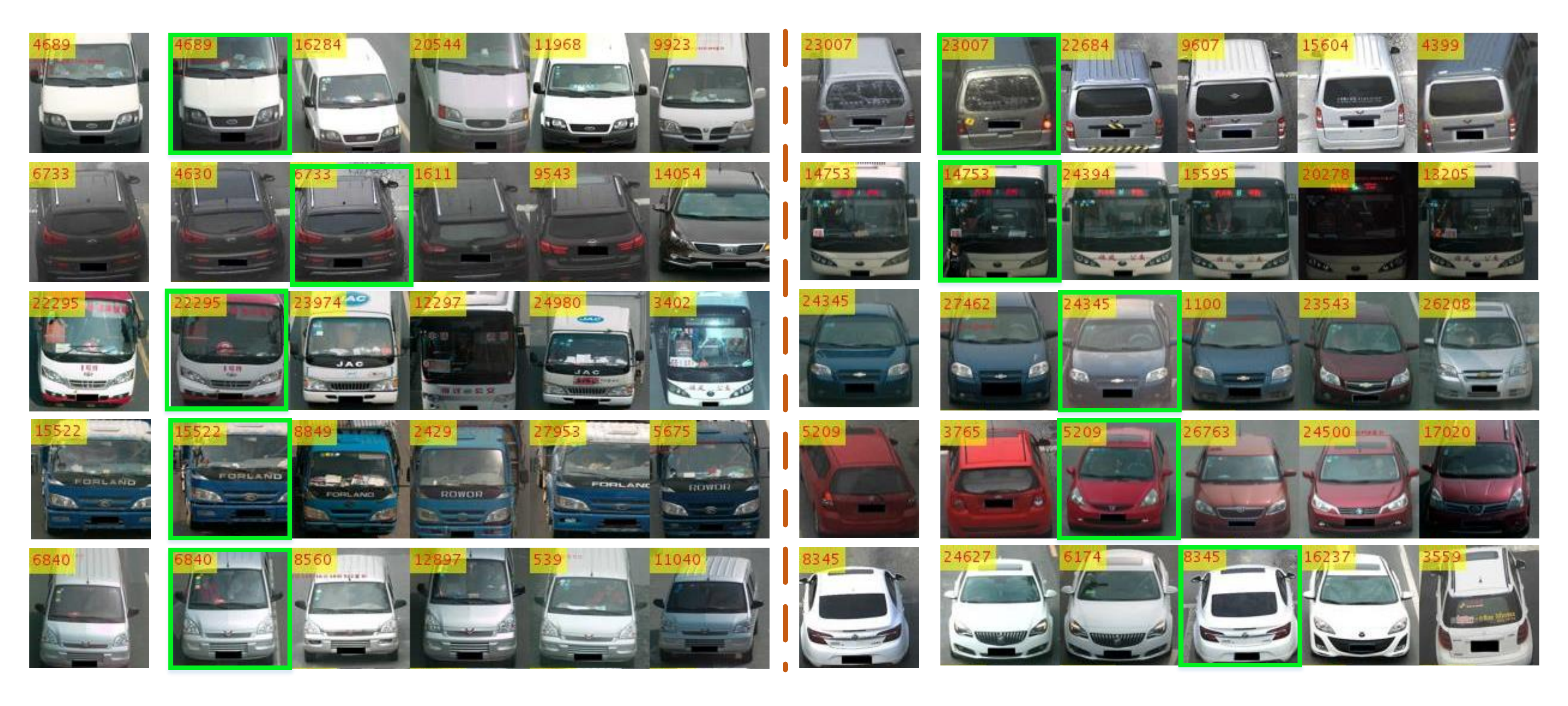}}
}
\caption{The retrieval results on the VehicleID and VeRi-776. (a) The results on VeRi-776. (b) The results on VehicleID. The left column shows query images while the images of right-hand side are retrieval results obtained by proposed method} \label{fig10}
\end{figure*}

Third, to demonstrate the effectiveness of the proposed ASL loss, ASL loss is compared with the ``AGNet-CE-All'' that means trains the reID model with cross-entropy loss and the same structure as ``AGNet-ASL-All''. Compared with ``AGNet-CE-All''. The significant improvements have achieved on both VeRi-776 and VehicleID, which verifies that the proposed ASL loss is more adaptive for the vehicle reID task. This is because that the ASL considers the vehicles with the same model and color should have high degree of similarity, even though they have different IDs. So it gives different weights for different conditions which could better regularize the network to train the vehicle reID model.

\subsection{Qualitative analysis}

To better illustrate that vehicle attribute is effective for the vehicle reID, we visualize the features of selected vehicle images in the VeRi-776 test set and project the features to 2-dimensional space. Then the t-SNE \cite{van2014accelerating} is employed for dimension reduction and visualization. We show all the vehicle test images in a non-occlusion form. As show in Fig.\ref{tsne}, it can be seen that not only the same identity can be clustered together, the vehicle images which have the same color or type are also clustered together, which demonstrates that the color and type are major cues for the vehicle reID.

To further illustrate the effectiveness of the proposed framework in this paper, some results are visualized. Examples of vehicle reID results on VeRi-776 and VehicleID by our approach are shown in Fig.\ref{fig10}. In Fig.\ref{fig10}, for VeRi-776, the left column shows query images, while images on the right-hand are the top-11 results obtained by the algorithm. Vehicle images with red border are error results while other images are right results. For VehicleID, the left column shows query images, while images on the right-hand side are the top-5 results obtained by the algorithm. Vehicle images with green border are right results while other images are wrong results. The number on the left-top means Vehicle ID/Camera ID. The same Vehicle ID represents the same vehicle. The Camera ID is the camera number that images are captured. From Fig.\ref{fig10}, it could be observed that we could observe that our proposed method has high accuracy and good robustness to different viewpoints and illumination.

\section{Conclusion}

In this paper, we propose AGNet with attribute-guided attention module which could learn global representation with the abundant attribute features in an end-to-end manner. Besides that, to better train the reID model, the ALS loss is presented, which can strength the distinct ability of vehicle reID model according to the attributes to regularize AGNet model. It can be observed from the results that compared with other existing vehicle reID methods, AGNet could achieve competitive results. However, it is difficult to distinct some vehicles that are occluded by other vehicles. Hence, in our future studies, we would focus on exploiting the local regions with distinct features to improves the performance of reID model.


%
%

\ifCLASSOPTIONcaptionsoff
  \newpage
\fi




\bibliographystyle{IEEEtran}
\bibliography{mybibfile}

\end{document}